\documentclass[11pt,a4paper]{article}
\usepackage[hyperref]{acl2021}
\usepackage{multirow}
\usepackage{tabularx}
\usepackage{float}
\usepackage{graphicx}
\usepackage{times}
\usepackage{latexsym}
\usepackage{amsmath}
\usepackage[ruled,vlined]{algorithm2e}
\usepackage[size=small]{todonotes}
\usepackage{booktabs}
\usepackage{enumitem}

\newcommand{\soli}[1]{\texttt{solidarity}}
\newcommand{\anti}[1]{\texttt{anti-solidarity}}
\newcommand{\rest}[1]{\texttt{other}}

\usepackage{microtype}

\aclfinalcopy 
\def\aclpaperid{831} 

\newcommand{\se}[1]{\textcolor{black}{#1}}
\newcommand{\senew}[1]{\textcolor{black}{#1}}
\newcommand{\ai}[1]{\textcolor{black}{#1}}
\newcommand{\ainew}[1]{\textcolor{black}{#1}}

\title{Changes in European Solidarity 
Before and 
During COVID-19:\\ Evidence from a Large Crowd- and Expert-Annotated Twitter Dataset}

\author{Alexandra Ils$^{\dagger}$, Dan Liu$^{\ddagger}$, Daniela Grunow$^{\dagger}$, Steffen Eger$^{\ddagger}$ \\
  $^{\dagger}$ Department of Social Sciences, Goethe University Frankfurt am Main \\
  $^{\ddagger}$ Natural Language Learning Group, Technische Universität Darmstadt \\
  \texttt{\{ils,grunow\}@soz.uni-frankfurt.de},\\
  \texttt{dan.liu.19@stud.tu-darmstadt.de}, 
  \texttt{eger@aiphes.tu-darmstadt.de}
  }

\date{}

\begin{document}
\maketitle

\begin{abstract}
We introduce the well-established social scientific concept of social solidarity and its contestation, anti-solidarity, as a new problem setting to supervised machine learning in NLP to assess how European solidarity discourses changed before and after 
the COVID-19 outbreak was declared a global pandemic. 
To this end, we annotate 
2.3k English and German tweets for (anti-)solidarity expressions, utilizing multiple human annotators and two annotation approaches (experts vs.\ crowds). 
We use these annotations to train 
a BERT model with multiple data augmentation strategies. 
Our augmented BERT model that combines both expert and crowd annotations outperforms the baseline BERT classifier trained with expert annotations only by 
over 25 points, from 58\% macro-F1 to almost 85\%. We use this high-quality model 
to automatically label over 
270k tweets between September 2019 and December 2020. 
We then assess the automatically labeled data for how statements related to European (anti-)solidarity discourses developed over time and in relation to one another, before and during the COVID-19 crisis. 
Our results show that solidarity became increasingly salient and contested 
during the crisis. 
While the number of solidarity tweets remained on a higher level and dominated the discourse in the scrutinized time frame,  anti-solidarity tweets initially spiked, 
then decreased to (almost)  pre-COVID-19 values before rising to a stable higher level until the end of 2020.
\end{abstract}

\section{Introduction}
Social solidarity statements and other forms of collective pro-social behavior expressed in online media have been argued to affect public opinion and political mobilization \cite{Fenton.2008, Margolin.2018, Santhanam.2019, Tufekci.2014}. 
The ubiquity of social media enables individuals to feel and relate to real-world problems through solidarity statements expressed online and to act accordingly \cite{Fenton.2008}. 
Social solidarity is a key feature that keeps modern societies integrated, functioning and cohesive. 
It constitutes a moral and normative bond between individuals and society, affecting people's willingness to help others and share own resources beyond immediate rational individually-, group- or class-based interests \cite{Silver.1994}. National and international crises intensify the need for social solidarity, 
as crises diminish the resources available, 
raise demand for new and additional resources, and/or require readjustment of established collective redistributive patterns, e.g.\ 
inclusion of new groups. Because principles of inclusion and redistribution are contested in modern societies and related opinions fragmented \cite{Fenton.2008, Sunstein.2018}, collective expressions of social solidarity online are likely contested. Such statements, which we refer to as anti-solidarity, question calls for social solidarity and its framing, i.e.\ towards whom individuals should show solidarity, and in 
\ainew{what} ways \cite{Wallaschek.2019}.

For a long time, social solidarity was considered to be confined to local, national or cultural groups. The concept of a European society and European solidarity \cite{Gerhards.2019}, a form of solidarity that goes beyond the nation state, is rather new. European solidarity gained relevance with the rise and expansion of the European Union (EU) and its legislative and administrative power vis-à-vis the EU member states since the 1950s \cite{Baglioni.2019, Gerhards.2019, koos.2019, Lahusen.2018}. After decades of increasing European integration and institutionalization, the EU entered into a continued succession of deep crises, beginning with the European \ainew{F}inancial \ainew{C}risis in 2010 \cite{Gerhards.2019}. Experiences of recurring European crises raise concerns regarding the future of European society and its foundation, European solidarity. Eurosceptics and right-wing populists claim that social solidarity is, and should be, confined within the nation state, whereas supporters of the European project see European solidarity as a means to overcome the great challenges imposed on EU countries and its citizens today \cite{Gerhards.2019}. To date, it is an open empirical question how strong and contested social solidarity really is in Europe, and how it has changed since the onset of the COVID-19 pandemic. Against this background, we ask whether we can detect changes in the debates on European solidarity before and after the outbreak of COVID-19. 
Our 
contributions are:
\begin{itemize}[topsep=5pt,itemsep=0pt,leftmargin=*]
 \item[(i)] We provide 
 a novel Twitter corpus 
 annotated for expressions of social solidarity and anti-solidarity. Our corpus contains 
 2.3k human-labeled tweets 
 from 
 two annotation strategies (experts vs.\ crowds). Moreover, we provide over 270k automatically labeled tweets based on an ensemble of BERT classifiers trained on the expert and crowd annotations. 
 \item[(ii)] 
 We train BERT on crowd- and expert annotations using multiple data augmentation and transfer learning approaches, achieving over 25 points 
 improvement over BERT trained on expert annotations alone. 
 \item[(iii)] We present novel empirical evidence regarding changes in European solidarity debates before and after the outbreak of the COVID-19 pandemic. Our findings show that both expressed solidarity and anti-solidarity escalated with the occurrence of incisive political events, such as the onset of the first European lockdowns. 
 \end{itemize}
 Our data and code are available from
 \url{https://github.com/lalashiwoya/socialsolidarityCOVID19}. 

\section{Related work}\label{sec:related}

\paragraph{Social Solidarity in the Social Sciences.} In the social sciences, social solidarity has always been a key topic of intellectual thought and empirical investigation, dating back to seminal thinkers such as Rousseau and Durkheim \cite{Silver.1994}. 
Whereas earlier \ainew{empirical} research 
was mostly confined to survey-based \cite{Baglioni.2019,Gerhards.2019,koos.2019,Lahusen.2018} or qualitative approaches \cite{Franceschelli.2019,GomezGarrido.2018,Heimann.2019}, 
computational social science just started tackling concepts as complex as solidarity as part of natural language processing (NLP) approaches \cite{Santhanam.2019}.

In (computational) social science, several studies investigated the European \ainew{M}igration \ainew{C}risis and/or the \ainew{F}inancial \ainew{C}risis as displayed in media discourses. These studies focused on differences in perspectives and narratives between mainstream media and Twitter, using topic models \cite{Nerghes.2019}, and the coverage and kinds of solidarity addressed in leftist and conservative newspaper media \cite{Wallaschek.2019, Wallaschek.2020}, as well as relevant actors in discourses on solidarity, using discourse network measures \cite{Wallaschek.2020b}. While these studies offer insight into solidarity discourses during crises, they all share a strong focus on mainstream media, which is unlikely to publicly reject solidarity claims \cite{Wallaschek.2019}. Social media, in contrast, allows its users to perpetuate, challenge and open new perspectives on mainstream narratives \cite{Nerghes.2019}. A first attempt to study solidarity expressed by social media users during crises has been presented by \citet{Santhanam.2019}. They assessed how emojis are used in tweets expressing solidarity relating to two crises
through hashtag-based manual annotation---\se{ignoring actual content of the tweets}---and utilizing a LSTM network for automatic classification. 
Their approach, while insightful, provides a rather simple operationalization of solidarity, which neglects its contested, consequential and obligatory aspects vis-à-vis other social groups. 

The current state of social science research on European social solidarity poses a puzzle. On the one hand, most survey research paints a rather optimistic view regarding social solidarity in the EU, despite marked cross-national variation \cite{Binner.2019, Dragolov.2016, Gerhards.2019, Lahusen.2018}. On the other hand, the rise of political polarization and Eurosceptic political parties \cite{Baker.2020, nicoli.2017} suggests that the opinions, orientations and fears of a potentially growing political minority is underrepresented in this research. People holding extreme opinions have been found to be reluctant to participate in surveys and adopt their survey-responses to social norms (social desirability bias) \cite{BazoVienrich.2017, Heerwegh.2009, Janus.2010}. Research indicates that such minorities may grow in times of crises, with both short-term and long-term effects for public opinion and political trust \cite{gangl.2018, nicoli.2017}. Our paper addresses these problems by drawing on large volumes of longitudinal social media data 
that reflect potential fragmentation of political opinion \cite{Sunstein.2018} and its change over time. Our approach will thus uncover how contested European solidarity is and how it developed since the onset of COVID-19. 

\paragraph{Emotion and Sentiment Classification in NLP.} \se{In NLP, annotating and classifying text (in social media) 
for sentiment or emotions is a 
well-established task \citep{demszky-etal-2020-goemotions, ding-etal-2020-hashtags, haider-etal-2020-po,  Hutto2014VADERAP, oberlander2018analysis}. 
Importantly, 
our 
approach focuses on 
\ai{expressions of} (anti-)solidarity: 
For example, texts containing a positive sentiment 
towards 
persons, groups or organizations which are at their core anti-European, nationalistic and excluding reflect anti-solidarity and are \ainew{annotated}
as such. 
Our annotations therefore 
go beyond superficial assessment of sentiment. \ai{In fact,} \se{
the correlation between sentiment labels---e.g., as 
obtained from 
Vader \citep{Hutto2014VADERAP}---and our 
annotations in \S\ref{sec:data} is only $\sim$0.2. 
\ai{Specifically,}
many tweets labeled as solidarity use negatively connoted emotion words.
}
} 
\section{Data and Annotations}\label{sec:data}
We use the unforeseen onset of the COVID-19 crisis, beginning with the first European lockdown, enacted late February to early March 2020\ainew{,}  
to analyze and compare social solidarity data before and during the COVID-19 crisis as if it were a natural experiment
\cite{Creighton.2015, Kuntz.2017}. In order to utilize this strategy and keep the baseline solidarity debate comparable before and after the onset of the COVID-19 crisis, we confined our sample to tweets with hashtags predominantly relating to two previous European crises whose effects continue to concern Europe, its member states and citizens: \textbf{(i)} Migration and the distribution of refugees among European member states, and \textbf{(ii)} Financial solidarity, i.e.\ financial support for indebted EU countries. The former solidarity debate predominantly refers to the Refugee Crisis since 2015 and the living situation of migrants, the latter mostly relates to the Financial Crisis, followed by the Euro Crisis, and concerns the excessive indebtedness of some EU countries since 2010.\footnote{Further analyses (not shown) revealed that around 20 percent of the tweets in our sample relate to solidarity regarding other issues.} 

\paragraph{Data.} We crawled 
271,930 \ai{tweets} 
between 01.09.2019 
\ainew{and} 31.12.2020, 
\ainew{written} in English or German \ainew{and} geographically restricted to 
Europe, to obtain setups comparable to the survey-based social science literature on European solidarity.
We only crawled tweets that contained specific hashtags, 
to filter for our two topics, i.e.\ refugee and 
financial solidarity. 
\ainew{We} started 
with an initial list of hashtags (e.g., 
``\#refugeecrisis'', ``\#eurobonds''), which we then expanded via co-occurrence statistics. \ainew{We manually evaluated 456 co-occurring hashtags with at least 100 occurrences to see if they \ainew{represented}
the topics we are interested in. 
Ultimately, we selected 45 hashtags (see appendix) to 
\ainew{capture} a \ainew{wide} range
{of} the discourse on migration and financial solidarity. 
Importantly, we keep the hashtag list associated with our 270k tweets constant over time.}\footnote{We follow a purposeful sampling frame, 
but this necessarily introduces a bias in our data. 
While we took care of including a variety of hashtags, we do not claim to have captured the full extent of discourse concerning the topics migration and financial solidarity.} 

\paragraph{Definition of Social Solidarity.} 
In line with social scientific concepts of social solidarity, we define social solidarity \ainew{as expressed and/or called for in online media} as “the preparedness to share one’s own resources with others, be that directly by donating money or time in support of others or indirectly by supporting the state to reallocate and redistribute some of the funds gathered through taxes or contributions” \citetext{\citealp[p.~4]{Lahusen.2018}}. 
We define anti-solidarity 
\ainew{as} expressions 
that contest this type of social solidarity and/or deny solidarity towards vulnerable social groups and other European 
states, e.g. by promoting nationalism or the closure of national borders \cite{Burgoon.Rooduijn.2021, Cinalli.2020, Finseraas.2008, Wallaschek.2017}. 

\paragraph{Expert Annotations.} 
After crawling and preparing the data, we set up guidelines for annotating tweets. 
Overall, we set four categories to annotate, with solidarity and anti-solidarity being the most important ones. A tweet indicating 
support for 
people in need, 
the willingness \ainew{and/or gratitude towards others} to share resources and/or help them is considered expressing \soli{}. 
The same applies to tweets \ainew{criticizing the EU in terms of not doing enough to share resources and/or help socially vulnerable groups as well as advocating for the EU as a solidarity union.} 
A tweet is considered \ai{to be expressing} \anti{} 
\ai{statements} if the above-mentioned criteria are reversed, \ainew{and/}or, the tweet contains tendencies of nationalism or advocates for closed borders. Not all tweets fit into these classes, 
thus we introduce two additional categories:
 \texttt{ambivalent} and \texttt{not applicable}. While the ambivalent category refers to tweets that could be interpreted as 
 both \ai{expressing} solidarity and anti-solidarity \ai{statements}, the second category is reserved for tweets that do not contain the topic of (anti-)solidarity at all \ainew{or refer to topics that are not concerned with discourses on refugee or financial solidarity}. Table \ref{table:examples} contains example tweets for all categories. Full guidelines for the annotation of tweets are given in the appendix.
 
We divided the annotation process into six working stages (I-VI) to refine our data set and annotation standards over time and strengthen inter-annotator reliability through subsequent discussions among annotators \ainew{and social science experts}. \se{Our annotators included four university students majoring in computer science, one computer science faculty member as well as two social science experts (one PhD student 
and one professor).} 
We started the training of \ainew{seven} annotators with a small dataset that 
\ainew{they annotated} independently and 
refined \ainew{the guidelines} during the annotation process. 
\ai{In} the training period, which lasted three iterations (I-III), we achieved Cohen's kappa values \ainew{of 0.51} among seven annotators. 
In working stage IV, 
two groups of two annotators 
annotated 
339 tweets with hashtags 
not included before. 
Across the four annotators, Cohen's kappa values \ainew{of 0.49} were reached. 
In working stages V and VI, 
one group of two 
\ainew{students annotated overall} 588 tweets, with a resulting kappa value of \ainew{0.79 and} 0.77 \ainew{respectively}. 

\ainew{While the kappa value was low in the first stages, we managed to raise the inter-annotator reliability over time through discussions with the social science experts and extension of the guidelines.} 
We also 
introduced a gold-standard for annotations from stage II onward \senew{which served as orientation}. 
This 
was determined by majority voting and discussions among the annotators. For cases where a decision on the gold-standard label could not be reached, 
\ainew{a social science expert} decided on the gold-standard label; some hard cases were 
left undecided (not included in the dataset).  

The gold-standard additionally served as human reference performance which we compared the model against. On average across all stages, our kappa agreement 
is 
\ainew{0.64} for four and 0.69 for three classes (collapsing \texttt{ambivalent} and \texttt{not applicable}), while the macro F1-score 
is 
\ainew{69\%} for four and 
78.5\% for three classes. 
However, in the final stages, the agreement is considerably higher: 
above 80\% macro-F1 for four and between 85.4\% and 89.7\% macro-F1 for three classes. 

\begin{table*}[!htb]
    \centering
    \begin{tabular}{l|l}
     \toprule
         Children caught up in the Moria camp fire face unimaginable horrors & \soli{} 
         \\
         \#SafePassage 
         \#RefugeesWelcome & \\ \midrule
         Most people supporting \#RefugeesWelcome are 
         racists or psychopaths &
         \anti{} \\ \midrule
         Does this rule apply to every UK citizen as well as every \#AsylumSeeker?  & \texttt{ambivalent} \\ \midrule
         Let's make \#VaccinesWork for everyone 
         \#LeaveNoOneBehind
         & \texttt{not applicable} \\
         \bottomrule
    \end{tabular}
    \caption{Paraphrased (and translated) sample of annotated tweets in our dataset, together with labels.}
    \label{table:examples}
\end{table*}

\begin{table}[!htb]
    \centering
    \begin{tabular}{c|cccc|c}
     \toprule
         &  S & A & AMB & NA & Total\\ \midrule
         Experts & 386 & 246 & 113 & 174 & 919\\
         Crowds & 768 & 209 & 186 & 217 & 1380\\
         \bottomrule
    \end{tabular}
    \caption{Number of annotated tweets (after geofiltering) for the four classes \soli{} (S), \anti{} (A), \texttt{ambivalent} (AMB), and \texttt{not applicable} (NA). }
    \label{table:statistics}
\end{table}

\paragraph{Crowd annotations.} We also conducted a `crowd experiment' with students in an introductory course to NLP. We provided students with the 
guidelines and 100 expert annotated tweets as illustrations. We trained crowd annotators in three iterations. 1) They were assigned reading the guidelines and 
\ainew{looking} at 30 random expert annotations\ainew{.}
\ainew{Then they were asked to annotate 20 tweets themselves} and self-report their kappa agreement with the experts (we provided the labels separately so that they could further use the 20 tweets to understand the annotation task). 2) We repeated this with another 30 tweets for annotator training and 20 tweets for annotator testing. 3) They received 30 \ainew{expert-annotated tweets for which we did not give them access to expert labels, }
and 30 entirely novel tweets, 
\ainew{that} had not been annotated before. These 60 final tweets were presented in random order to each student. \se{50\% of the 30 novel tweets were taken from before September 2020 and the other 50\% were taken from after September 2020.}

 125 students participated in the annotation task. 
 \ai{The} annotation experiment was part of a bonus the students could achieve for the course (counted 12.5\% of the overall bonus for the class). Each novel tweet was annotated by up to 3 students (2.7 on average). 
 To obtain a unique label for each crowd-annotated tweet, we used the following simple strategy: we either chose the majority label among the three annotators or the annotation of the most reliable annotator in case there was no unique majority label. The \ainew{annotator that had the highest agreement with the expert annotators was taken as most reliable annotator.} 

\begin{figure}[htb]
    \centering
    \includegraphics[scale=0.5]{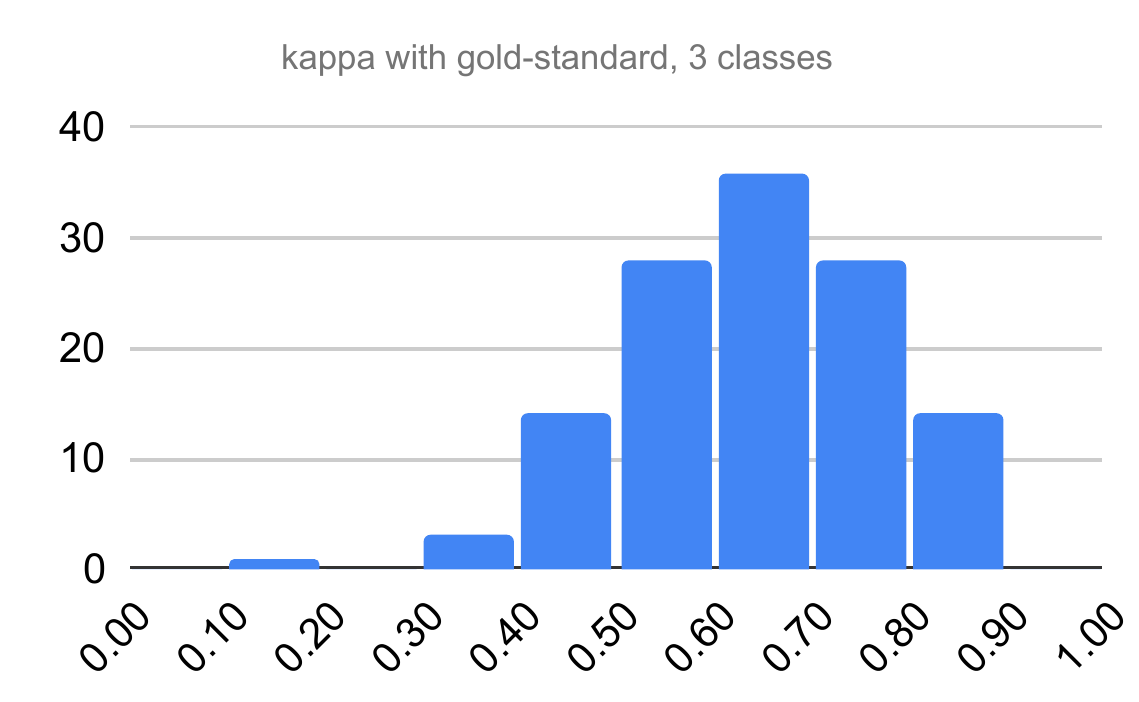}
    \caption{Distribution of kappa agreements of crowd workers with expert annotated gold-standard, 3 classes.}
    \label{fig:crowdvsexperts}
\end{figure}

Kappa agreements of students with the experts 
are shown in Figure \ref{fig:crowdvsexperts}. 
The majority of students has a kappa agreement with the gold-standard of between 0.6-0.7 
when three classes are taken into account 
and between 0.5-0.6 
for four classes. 

In Table \ref{table:statistics}, we further show statistics on our annotated datasets: we have 2299 
annotated tweets in total, 
about 60\% of which have been annotated by crowd-workers. About 50\% of all tweets are annotated as 
\soli{}, 
20\% as 
\anti{}, 
and 30\% as either \texttt{not-applicable} or \texttt{ambivalent}. 
In our annotations, 
1196 tweets are English 
and 1103 
are German.\footnote{\se{In our automatically labeled data, the majority of tweets is German. We assumed all German tweets to come from within the EU, while the English tweets would be geofiltered more aggressively.}}
\se{Finally, we note that the distribution of labels for expert and crowd annotations are different, i.e., the crowd annotations cover more solidarity tweets. The reason is twofold: (a) for the experts, we oversampled hashtags that we believed to be associated more often with anti-solidarity tweets as the initial annotations indicated that these would be in the minority, which we feared to be problematic for the automatic classifiers. (b) The time periods in which the tweets for the experts and crowd annotators fall differ.}
\section{Methods}
We use multilingual BERT \citep{bert} / XLM-R \citep{conneau-etal-2020-unsupervised} to classify our tweets in a 3-way 
classification problem (\soli{}, \anti{}, \rest{}), \se{not differentiating between the classes \texttt{ambivalent} and \texttt{non-applicable} since our main focus is on the analysis of changes in (anti-)solidarity}. 
We use the baseline MBERT model: bert-base-multilingual-cased and the base XLM-R model: xlm-roberta-base. 
We implemented several data augmentation/transfer learning techniques to improve model performance: 
\begin{itemize}[topsep=5pt,itemsep=0pt,leftmargin=*]
    \item \textbf{Oversampling of minority classes}: We randomly duplicate (expert and crowd annotated) tweets from minority classes until all 
    classes have the same number of tweets as the majority class \soli{}. 
    \item \textbf{Back-translation}: We use the Google Translate API to translate English tweets into a pivot language (we used German), and pivot language tweets back into English (for expert and crowd-annotated tweets). 
    \item \textbf{Fine-tuning}: We 
    fine-tune MBERT / XLM-R with masked language model and next sentence prediction tasks on domain-specific data, i.e., our crawled unlabeled tweets.
    \item \textbf{Auto-labeled data}: 
    As a form of self-learning, we 
    train 9 different models (including oversampling, back-translation, etc.) on the expert and crowd-annotated data, then apply them to our full dataset (of 270k tweets, see below). We only retain tweets where 7 of 9 models agree and select 35k such tweets for each label 
    (\soli{}, \anti{}, \rest{})
    into an augmented training set, thus increasing training data by 105k auto-labeled tweets.  
    \item \textbf{Ensembling}: We take the majority vote of 15 different models 
    to  
    leverage heterogeneous information.
    The $k=$ 15 models, like the $k=$ 9 models above, were determined as the top-$k$ models by their dev set performance. 
\end{itemize}
We also experimented with re-mapping multilingual BERT and XLM-R \citep{cao2020multilingual,zhao2020inducing,zhao-etal-2020-limitations} as they have not seen parallel data during training, but found only minor effects in initial experiments. 
\section{Experiments}

\begin{table*}[!htb]
    \centering
    \begin{tabular}{l|r|ll|ll}
    \toprule
                &               & \multicolumn{2}{c|}{MBERT} & \multicolumn{2}{c}{XLM-R}\\
         \multicolumn{1}{c|}{\textbf{Condition}} & \textbf{Train size} & \multicolumn{1}{c}{\textbf{Dev}} & \multicolumn{1}{c|}{\textbf{Test}} &
         \multicolumn{1}{c}{\textbf{Dev}} & \multicolumn{1}{c}{\textbf{Test}}
         \\ \midrule
         E, Hashtag only & 572 & 51.7$\pm$0.5 & 49.0$\pm$1.1 & 48.0$\pm$0.9 & 44.0$\pm$0.8\\
         E & 579 & 64.2$\pm$1.2 & 57.7$\pm$0.4 & 64.0 & 63.3\\
         E+C & 1959 & 66.4$\pm$0.5 & 64.0$\pm$1.5 & 65.0 &	64.8\\
         E+C, No Hashtags & 1959 & 64.0$\pm$0.3 & 58.0$\pm$0.5 & 62.0 & 60.0\\
         E+C, Hashtag Only & 1567 & 55.8$\pm$2.0 & 49.5$\pm$2.1 & 47.8 & 42.2\\
         E+C+Auto label & 106959 & 76.4 & 78.3 & 77.5 & 78.4\\ 
         E+C+Auto label+Oversample & 108048 & 76.4 & 76.3 & 77.4 & 76.9 \\
         E+C+Auto label+Backtranslation & 108918 & 76.0 & 77.1 & 77.5 & 78.7 \\
         E+C+Auto label+Pretraining & 106959 & 78.4 & 78.8 & 78.6 & 79.0 \\ 
         E+C+ALL & 110007 & 78.8$\pm$1.3 & 78.6$\pm$0.8 & 78.9 & 79.7 \\
    \bottomrule
    \end{tabular}
    \caption{Macro-F1 scores (in \%) for different conditions. Entries with $\pm$ give averages and standard deviations over 3 different runs with different test and dev sets. `E' stands for Experts, `C' for Crowds. `ALL' refers to all data augmentation and transfer learning techniques.}
    \label{table:full}
\end{table*}

In \S\ref{sec:setup}, we describe our experimental setup. 
In \S\ref{sec:solidarity}, we show the classification results of our baseline models on the annotated data and the effects of our various data augmentation and transfer learning strategies. 
In \S\ref{sec:model_analysis}, we analyze performances of our best-performing models. 
In \S\ref{sec:temporal}, we automatically label our whole dataset of 270k tweets and analyze changes in solidarity over time. 
 
\subsection{Experimental Setup}\label{sec:setup}
To examine the effects of various factors, we design several experimental conditions. These involve (i) using only hashtags for classification, ignoring the actual tweet text, 
(ii) using only text, without the hashtags, 
(iii) combining expert and crowd annotations for training, (iv) examining the augmentation and transfer learning strategies, (v) ensembling various models using majority voting. 

All models are evaluated on randomly sampled test and dev sets of size 170 each. Both dev and test set are taken from the expert annotations.  
\se{We use the dev set for early stopping.}
To make sure our results are not an artefact of unlucky choices of test and dev sets, we report averages of 3 random splits where test and dev set contain 170 instances in each case (for reasons of computational costs, we do so only for selected experimental conditions). 

We report the macro-F1 score to evaluate the performance of different models. \se{Hyperparameters of our models can be found in our github}.

\subsection{Results}\label{sec:solidarity}
The main results are reported in Table \ref{table:full}. 
Using only hashtags and expert annotated data yields a macro-F1 score of below 50\% for MBERT and XLM-R. Including the full texts improves this by over 8 points (almost 20 points for XLM-R). 
\ainew{Adding} crowd-annotations yields another substantial boost of more than 6 points for MBERT. Removing hashtags in this situation decreases the performance between 5 and 6 points. This means that the hashtags indeed contain import information,  
but the texts are more important than the hashtags: with hashtags only, we observe macro-F1 scores between 42 and 49\%, whereas with text only the performances are substantially higher, between 58 and 60\%. While using hashtags only means less data since not all of our tweets have hashtags, the performance with only hashtags on the test sets 
\ainew{stays} below 50\%, both with 572 and more than 1500 tweets for training. 

Next, we analyze the data augmentation and transfer learning techniques. Including auto-labeled data drastically increases the train set, from below 2k instances to over 100k. Even though these instances are self-labeled, performance increases by over 13 points to about 78\% macro-F1. Additionally oversampling or backtranslating the data does not yield further benefits, but pretraining on unlabeled tweets is effective even here and boosts performance to over 78\%. Combining all strategies yields scores of up to almost 80\%. Finally, when we consider our ensemble of 15 
models, 
we achieve 
a best performance of 84.5\% macro-F1 on the test set, close 
to the human macro-F1 agreement for the experts in the last rounds of annotation. 

\se{
To sum up, we note: 
(i) adding crowd annotated data 
clearly helps, 
despite the crowd annotated data having a different label distribution; (ii) including text is important for classification 
\ainew{as} 
the classification with hashtags only performs considerably worse; (iii) data augmentation (especially self-labeling), 
combining models and transfer learning strategies has a further clearly positive effect. 
}

\subsection{Model Analysis}\label{sec:model_analysis}
\se{Our most accurate ensemble models perform best for the majority class \soli{} with an F1-score of almost 90\%, about 10 points better than for \anti{} and over 5 points better than for the \rest{} class. 
}
A confusion matrix for this best performing model is shown in Table \ref{table:confusion}.  
Here, 
\anti{} is 
disproportionately misclassified as either 
\soli{} or the 
\rest{} class. 

\begin{figure*}[!htb]
    \centering
    \includegraphics[scale=0.5]{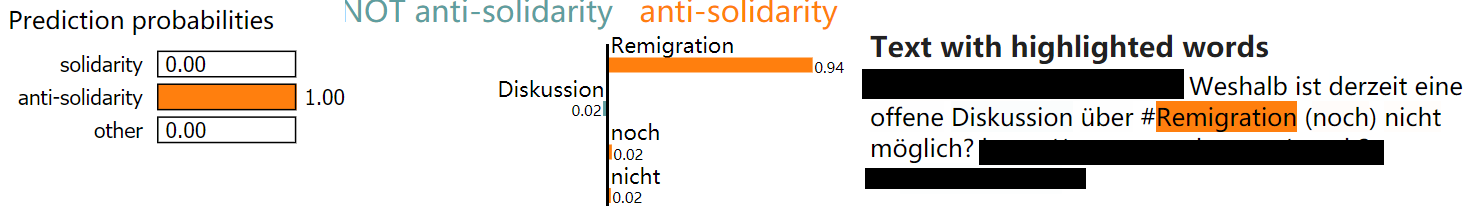}
    \caption{Our best-performing model (macro-F1 of 84.5\%) predicts 
    \anti{} 
    for the current example because of the hashtag \#Remigration (according to LIME). The tweet, also given as translation in Table \ref{table:misclassifications} (2) below, is overall classified as 
    \rest{} 
    in the gold standard, as it may be considered as expressing no determinate stance. Here, we hide identity revealing information in the tweet, but our classifier sees it.}
    \label{fig:lime}
\end{figure*}

Table \ref{table:misclassifications} shows selected misclassifications for our 
ensemble model with performance of about 84.5\% macro-F1. 
This reveals that the models 
sometimes leverage superficial lexical cues (e.g., 
\ainew{the German} political party `AfD' is typically associated with anti-solidarity towards EU and refugees), including 
hashtags (`Remigration'); see Figure \ref{fig:lime}, where we used LIME \cite{lime} to highlight words the model pays attention to.
To further gain insight into the misclassifications, we had one social science expert reannotate all misclassifications. 
From the 25 errors that our best model makes in the test set of 170 instances, the expert thinks that 
\ainew{12} times the gold standard is correct, 7 times the model prediction is correct, 
and in further 
\ainew{6} cases neither the model nor the gold standard are correct. This hints at some level of errors in our annotated data; 
it further supports the conclusion that our model is close to the human upper bound. 

\begin{table}[!htb]
\centering
\begin{tabular}{cc|ccc}
     \toprule
     & & \multicolumn{3}{c}{Predicted} \\ 
          & & \textbf{S} & \textbf{A} & \textbf{O} \\ \midrule
          & \textbf{S} & 63 & 3 & 2  \\
     Gold & \textbf{A} & 5 & 37 & 4 \\
          & \textbf{O} & 5 & 6 &  45 \\
     \bottomrule
\end{tabular}
\caption{Confusion matrix for best ensemble with macro-F1 score of 84.5\% on the test set (for one specific train, dev, test split). 
}
\label{table:confusion}
\end{table}
\begin{table*}[!htb]
\centering
\begin{tabular}{l|cc}
     \toprule
     \multicolumn{1}{c}{\textbf{Text}} & \textbf{Gold} & \textbf{Pred.} \\ \midrule
     (1) 
     You can drink a toast with 
     the AFD misanthropists \#seenotrettung \#NieMehrCDU & S & A \\
     (2) Why is an open discussion about \#Remigration (not) yet possible? & O & A \\
     \midrule
     (3) Raped and Beaten, Lesbian \#AsylumSeeker Faces \#Deportation & A & O 
     \\ \bottomrule
\end{tabular}
\caption{Selected misclassifications of best performing ensemble model. We consider the bottom tweet misclassified in the expert annotated data (correct would be \soli{}). Tweets are paraphrase\ainew{d}
and/or translated.}
\label{table:misclassifications}
\end{table*}

\subsection{Temporal Analysis}\label{sec:temporal}
\begin{figure*}[!htb]
    \centering
    \includegraphics[width=16cm, height=6.9cm]{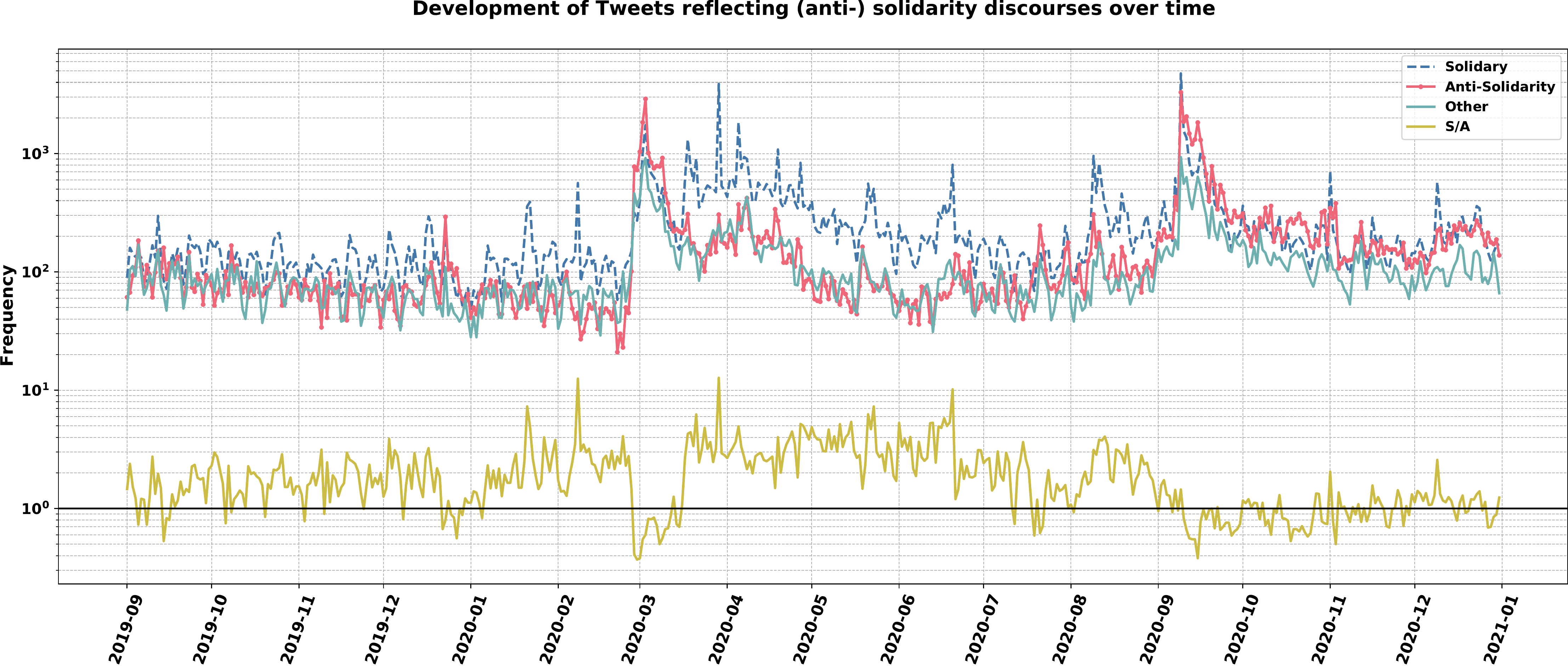}
    \caption{
    Frequency of \soli{} (S), \anti{} (A) and \rest{} \ainew{(O)} tweets over time as well as the ratio $S/A$. The constant line 1 indicates where $S=A$. Y-axis is in log-scale.
    }
    \label{fig:trend}
\end{figure*}

Throughout the period observed in our data, discourses relating to migration were much more frequent than financial solidarity discourses. We crawled an average of 2526 tweets per week relating to migration (anti-)solidarity and an average of 174 financial (anti-)solidarity tweets, \se{judging from the associated hashtags}. 

\se{We used our best performing model  
to automatically label all our 270k tweets between September 2019 and December 2020. 
Solidarity tweets were about twice 
as frequent compared to anti-solidarity tweets, reflecting a polarized discourse in which solidarity statements clearly dominated. 
Figure \ref{fig:trend} shows the frequency curves for \soli{}, \anti{} and \rest{} tweets over time in our sample. The figure also gives the ratio
\begin{align*}
    S/A \: := \frac{\#\text{Solidarity tweets}}{\#\text{Anti-Solidarity tweets}} 
\end{align*}
that shows the frequency of solidarity tweets relative to anti-solidarity tweets.
} Values above one indicate that more solidarity than anti-solidarity statements were tweeted that day. 

Figure \ref{fig:trend} displays several short-term increases in solidarity statements 
in our window of observation. Further analysis shows that these peaks have been immediate responses 
to drastic politically relevant events in Europe, which were also prominently covered by mainstream media, i.e.\ COVID-19-related news, natural disasters, fires, major policy changes. 
\se{We illustrate this in the following.}

On March 11th 2020, the World Health Organization (WHO) declared the COVID-19 outbreak a global pandemic. Shortly before and after, European countries started to take a variety of countermeasures, including stay-at-home orders for the general population, private gathering restrictions, and the closure of educational \ainew{ and childcare} institutions 
\cite{ECDC.2020a}. With the onset of these interventions, both solidarity and anti-solidarity statements relating to refugees and financial solidarity increased dramatically. 
At its peak at the beginning of March, anti-solidarity statements 
markedly 
outnumbered solidarity statements 
(we recorded 2189 solidarity tweets vs.\ 2569 anti-solidarity tweets on march 3rd). \se{In fact, the period in early March 2020 is the only extended period in our data where anti-solidarity statements outweighed solidarity statements.} 
The 
dominance of solidarity statements was reestablished after two weeks. Over the following months, anti-solidarity statements decreased again to pre-COVID-19 levels, whereas solidarity statements remained comparatively high, with several peaks between March and September 2020. 

Solidarity and anti-solidarity statements shot up again early-September 2020, with an unprecedented climax on September 9th. 
Introspection of our data shows that the 
trigger for this 
was 
the precarious situation of refugees after a fire destroyed the Mória Refugee Camp on the Greek island of Lesbos on the night of September 8th. Human Rights Watch had compared the camp to an open-air prison in which refugees lived under inhumane conditions, and the disaster spurred debates about the responsibilities of EU countries towards refugees and the countries hosting refugee hot spots (i.e.\ Greece and Italy). At \ainew{that} 
time, COVID-19 infection rates in the EU were \ainew{increasing} but 
still 
low,
and national measures to prevent the spread of infections relaxed in some and tightened in other EU countries \cite{ECDC.2020a, ECDC.2020b}. Further analyses (not displayed) show that the dominance of solidarity over anti-solidarity statements at the time was driven by tweets using hashtags relating to migration. The contemporaneous discourse on financial solidarity between EU countries was much less pronounced.  
\senew{From September 2020 to December 2020, solidarity and (anti-)solidarity statements were about equal in frequency, which means that anti-solidarity was on average on a higher level compared to the earlier time points in our time frame. This period also corresponds to the highest COVID-19 infection rates witnessed in the EU, on average, during the year 2020. In fact, the  Spearman correlation between the number of anti-solidarity tweets in our data and infection rates is 0.45 and 0.47, respectively (infection rates within  Germany \ainew{and}
the EU); \ainew{see Figure \ref{figure:infection} in the appendix.} 
Correlation with the number of  solidarity tweets is\ainew{, in contrast,} non-significant.
}
\paragraph{Discussion} 
Late February to mid-March 2020, EU governments began enacting lockdowns and other measures to contain COVID-19 infection rates, turning people’s everyday lives upside down. 
\senew{During this time frame, anti-solidarity statements peaked in our data, but solidarity statements quickly dominated thereafter again}. 
 During the summer of 2020, anti-solidarity tweets decreased whereas solidarity tweets continued to prevail on higher levels than before. 
 A major peak on September 9th, in the aftermath of the destruction of the Mória Refugee Camp, signifies an intensification of the polarized solidarity discourse. 
 \senew{From September to December 2020, anti-solidarity and solidarity statements were almost equal in number. Thus, 
 the onset of the COVID-19 crisis as well as times of high infection rates concurred with disproportionately high levels of anti-solidarity, despite a dominance of solidarity overall. Whether the relationship between anti-solidarity and intensified strains during crises is indeed causal will be the scope of our future research.}\footnote{We made sure that the substantial findings reported here are not driven by inherently German (anti-)solidarity discourses. Still, our results are bound to the opinions of people posting tweets in the English and German language.} 

\section{Conclusion}
\se{In this paper, we contributed the first large-scale human and automatically annotated dataset labeled for 
solidarity and its contestation, anti-solidarity. The dataset 
uses the textual material in social media posts to determine whether a post shows (anti-)solidarity with respect to relevant target groups. Our annotations, conducted by both trained experts and student crowd-workers, show overall good agreement levels for a challenging novel NLP task. We further trained augmented BERT models whose performance 
is close to the agreement levels of the experts and which we used for large-scale trend analysis of over 270k media posts before and after the onset of the COVID-19 pandemic. \ainew{Our findings 
show that (anti-)solidarity statements climaxed momentarily with the first lockdown, but the predominance of solidarity expressions was quickly restored at higher levels than before.}
\senew{Solidarity and anti-solidarity statements were balanced by the end of the year \ainew{2020}, when infection rates were rising.}
} 

The COVID-19 pandemic constitutes a worldwide crisis, with profound economic and social consequences for contemporary societies. It 
manifests yet another challenge for European solidarity, by putting a severe strain on available resources, i.e.\ national economies, health systems, and individual freedom. While the EU, its member countries and residents continued \ainew{to struggle} 
with the consequences of the  \ainew{F}inancial \ainew{C}risis and its aftermath, as well as migration, the COVID-19 pandemic \ainew{has} accelerated the problems related to these former crises. 
\senew{Our data suggests that }
the COVID-19 \ainew{pandemic}
\senew{has not severely negatively impacted }
\ainew{the willingness of} European Twitter users 
to take responsibility for refugees, while financial solidarity with 
other EU countries remained low on the agenda.  \ainew{Over time, however, this form of expressed solidarity became more controversial.}
\ainew{On one hand, t}hese findings are in line with survey-based, quantitative research and its rather optimistic overall picture regarding social solidarity in the EU during earlier crises \cite{Baglioni.2019, Gerhards.2019, koos.2019,Lahusen.2018}; \ainew{on the other hand, results from our correlation analysis suggests that severe strains during crises coincide with increased levels of anti-solidarity statements.}
We conclude
that a convergence of opinion \cite{Santhanam.2019} among the European Twitter-using public regarding the target audiences of solidarity, and the 
limits of European solidarity vs.\  national interests, is not in sight. Instead, our widened analytic focus \ainew{has allowed}
us to 
examine 
pro-social online behavior during crises 
and 
its opposition, \ainew{revealing}
that European Twitter users remain divided on issues 
of 
European solidarity.

\section*{Acknowledgments}
We thank the anonymous reviewers whose comments greatly improved the final version of the paper. 

\newpage

\paragraph{Ethical considerations.} We will release only tweet IDs in our final dataset. The presented tweets in our paper were paraphrased and/or translated and therefore cannot be traced back to the users. No user identities of any annotator (neither expert nor crowd worker) will ever be revealed or can be inferred from the dataset. Crowd workers were made aware that the annotations are going to be used in further downstream applications and they were free to choose to submit their annotations. While our trained model could potentially be misused, we do not foresee greater risks than with established NLP applications such as sentiment or emotion classification. 

\bibliography{acl2020}
\bibliographystyle{acl_natbib}

\clearpage
\appendix
\section{Appendices}
\section*{Guidelines}
Read the guidelines for annotating solidarity carefully.

\begin{itemize}
\item \underline{Definition of solidarity}: 
\begin{quote} The preparedness to share one’s own resources with others, be that directly by donating money or time in support of others or indirectly by supporting the state to reallocate and redistribute some of the funds gathered through taxes or contributions \cite{Lahusen.2018}
\end{quote}
\item General rules
\begin{enumerate}
    \item Do not take links (urls) into account when annotating.
    \item Hashtags should be taken into account, especially if a tweet is otherwise neutral.
    \item Emojis, if easily interpretable, can be taken into account. 
    \item If solidarity and anti-solidarity hashtags are used, code anti-solidarity.
    \item When annotating use the scheme: Solidarity: 0,
    Anti-Solidarity: 1,
    Ambivalent: 2,
    Not Applicable: 3. 
\end{enumerate}
\item Detailed rules for annotation. 
\begin{enumerate}
\item A tweet is annotated as showing \textbf{solidarity}, when:
\begin{enumerate}
    \item It clearly indicates support people and the willingness to share resources and/or help. 
    \item Positive attitude and gratitude to those sharing resources and/or helping. 
    \item  Advocacy of the European Union as a solidarity union.
    \item Criticism of the EU in terms of not doing enough to share resources and/or help others.
    \item Hashtags can be to be taken into account as to whether a tweet qualifies as showing solidarity (e.g.\ using hashtags like \texttt{\emph{\#refugeeswelcome}}).
    \item Hashtags should be taken into account if the tweet points neither towards solidarity or anti-solidarity itself.
    \newline

\end{enumerate}    

\item A tweet is annotated as showing \textbf{anti-solidarity}, when:
\begin{enumerate}
    \item It clearly indicates no willingness to support people and an unwillingness to share resources and/or help.
    \item It suggests to exclude our target groups from resources they currently have access to.
    \item Tendencies of nationalism and closing borders.
    \item Irony/sarcasm in tweets need to be taken into account.
    \item Hashtags can be taken into account as to whether a tweet qualifies as anti-solidarity (e.g.\ using hashtags like \texttt{\emph{\#grexit}}).
    \item Hashtags should be taken into account if the tweet points neither towards solidarity or anti-solidarity itself.
    \newline  
\end{enumerate}

\item A tweet is annotated \textbf{ambivalent}, when:
\begin{enumerate}
    \item The tweet shows solidarity or anti-solidarity sentiment, but it cannot be determined whether the tweet shows solidarity or anti-solidarity as there is additional info missing.
    \item Even if taking hashtags into account, there is no clear indication as to whether the author shows solidarity or anti-solidarity.
    \item If solidarity and anti-solidarity hashtags are used, code anti-solidarity.
    \newline 
\end{enumerate}

\item A tweet is annotated \textbf{not applicable}, when:
\begin{enumerate}
    \item There is no indication of solidarity or anti-solidarity sentiment in the tweet.
    \item Even when hashtags that usually point towards solidarity or anti-solidarity are taken into account the tweet does not indicate any connection to solidarity or anti-solidarity.
    \item The tweet concerns completely different topics than solidarity or anti-solidarity.
    \item The tweet is not understandable (e.g. contains only links).
    \newline 
\end{enumerate}
\end{enumerate}
\end{itemize}

\section*{Hashtags}
\begin{table}[!htb]
\begin{center}
\begin{tabular}{ |c|c| } 
 \hline
 Refugee crisis hashtags & Finance crisis hashtags  \\ 
 \hline
 \#asylumseeker & \#austerity + \#eu   \\ 
 \#asylumseekers & \#austerity + \#euro  \\ 
 \#asylkrise  &  \#austerity + \#eurobonds  \\ 
 \#asylrecht  & \#austerity + \#europe  \\ 
 \#asylverfahren & \#austerity + \#eurozone \\ 
 \#flüchtling   & \#austerität + \#eu  \\ 
 \#flüchtlinge  & \#austerität + \#euro  \\ 
 \#flüchtlingskrise &  \#austerität + \#eurobonds  \\
 \#flüchtlingswelle  &  \#austerität + \#europa  \\ 
 \#leavenoonebehind  & \#austerität + \#eurozone  \\ 
 \#migrationskrise  & \#debtunion   \\ 
 \#niewieder2015  & \#eurobonds  \\ 
 \#opentheborders  & \#eurocrisis   \\ 
 \#refugee & \#eurokrise  \\ 
 \#refugees & \#eusolidarity    \\ 
 \#refugeecrisis  & \#eusolidarität   \\ 
 \#refugeesnotwelcome  & \#exiteu   \\
 \#refugeeswelcome & \#fiscalunion \\
 \#rightofasylum   & \#fiskalunion   \\ 
 \#remigration & \#schuldenunion    \\ 
 \#seenotrettung  & \#transferunion   \\ 
 \#standwithrefugees  &  \\ 
 \#wirhabenplatz &  \\ 
 \#wirschaffendas &  \\ 
 \hline
\end{tabular}
\end{center}
\caption{Hashtags used in our experiments.}
\end{table}

\section*{Infection numbers vs.\ anti-solidarity Tweets}

\begin{figure*}[!htb]
    \centering
    \includegraphics[scale=0.5]{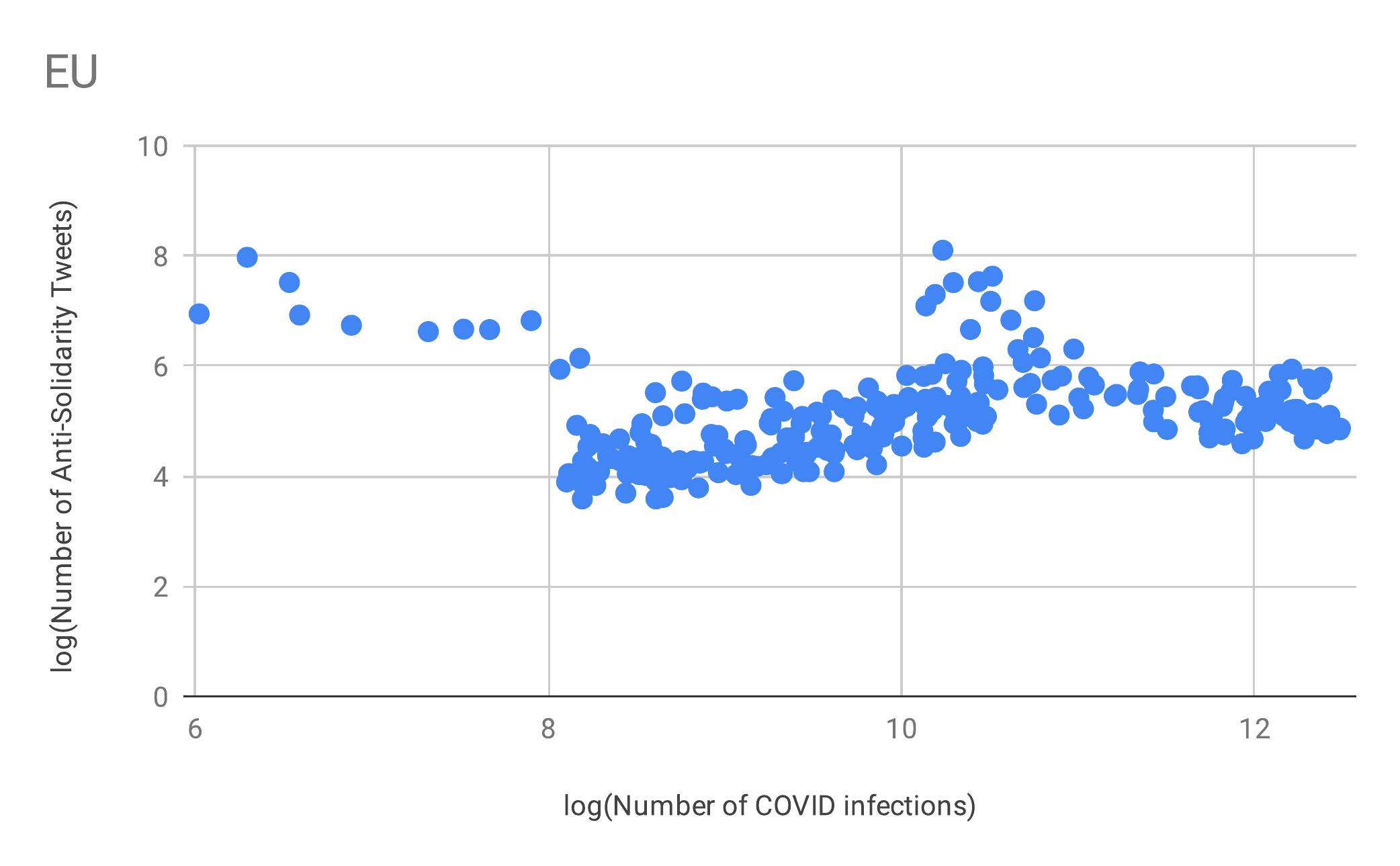}
    \includegraphics[scale=0.5]{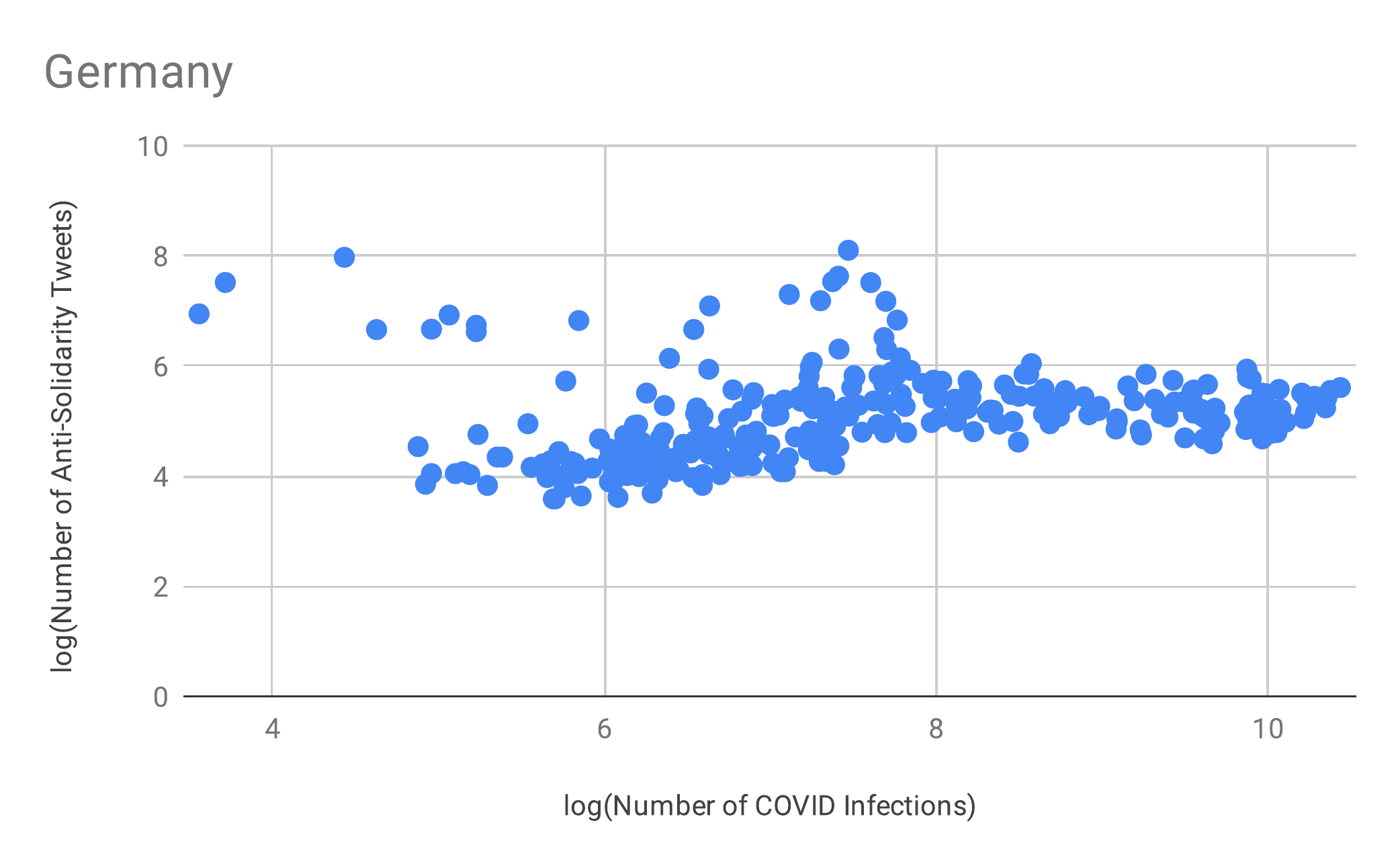}
    \caption{Scatter plot between infection rates and number of anti-solidarity tweets. Top: EU. Bottom: Germany. The time frame under consideration is 01.03.2020 to 14.12.2020 based on the data from \url{https://www.ecdc.europa.eu/en/publications-data/download-todays-data-geographic-distribution-covid-19-cases-worldwide}, restricted to 28 EU countries.}
    \label{figure:infection}
\end{figure*}

\end{document}